\documentclass[11pt,a4paper]{article}
\usepackage[hyperref]{acl2019}
\usepackage[utf8]{inputenc}
\usepackage{times}
\usepackage{latexsym}
\usepackage{graphicx}

\usepackage{url}

\aclfinalcopy 

\title{Let's FACE it. Finnish Poetry Generation with Aesthetics and Framing}

\author{Mika Hämäläinen \\
  Department of Digital Humanities \\
  University of Helsinki \\
  \texttt{mika.hamalainen@helsinki.fi} \\\And
  Khalid Alnajjar \\
  Department of Computer Science (HIIT) \\
  University of Helsinki \\
  \texttt{khalid.alnajjar@helsinki.fi} \\}

\date{}

\begin{document}
\maketitle
\begin{abstract}
  We present a creative poem generator for the morphologically rich Finnish language. Our method falls into the master-apprentice paradigm, where a computationally creative genetic algorithm teaches a BRNN model to generate poetry. We model several parts of poetic aesthetics in the fitness function of the genetic algorithm, such as sonic features, semantic coherence, imagery and metaphor. Furthermore, we justify the creativity of our method based on the FACE theory on computational creativity and take additional care in evaluating our system by automatic metrics for concepts together with human evaluation for aesthetics, framing and expressions.
\end{abstract}

\section{Introduction}

This paper explores the topic of computational creativity in the case of poem generation in Finnish. Our work does not only aim to generate, but rather create poems automatically. We take the FACE model \cite{colton2011computational} for computational creativity as our definition of creativity. Through this model, we motivate and evaluate creativity exhibited by our system.

Methodologically, our work embraces the master-apprentice method \cite{inlg} used in the past for computationally creative tasks. This means using a creative genetic algorithm as a master to teach an apprentice which is sequence-to-sequence neural network model. This way the overall system can approximate creative autonomy \cite{Jennings2010} if the apprentice was to be exposed to data originating from another source than the master. For further discussion on the topic of autonomy, see the original work establishing the master-apprentice method.

We pay special attention to evaluation of our system, and we motivate it through the FACE model. A creative system should be evaluated in terms of what has actually been modelled rather than on an ad-hoc and unjustified fashion. Additionally, our contribution lies on the fact that the aesthetics of our system are motivated by existing non-computational literature in poetry analysis. Furthermore, our system is capable of adjusting its aesthetics based on existing poetry.

Our work sheds some more light into the nature of a master-apprentice system. Especially by seeking to answer the question of multiple masters raised in the original work on the topic \cite{inlg}, which the authors left unanswered.

\section{Related Work}

While poetry generation has been tackled a number of times before by multiple authors \cite{gervas2001expert,jukka,misztal2014poetry,oliveira2017multilingual}, and an excellent overview is provided by \citet{goncalo-oliveira-2017-survey} on the recent state of the research, we dedicate this section in describing the most recent work conducted in the field after the aforementioned overview paper.

TwitSong \cite{twitsong3} mines a corpus for verses to be used in poetry based on how well they rhyme together. They score the verses in poems by four metrics (\textit{meter, emotion, topicality} and \textit{imagery)} and use a genetic algorithm to edit the worst scoring verse in the poem. However, they only assess poems on a verse level and their algorithm lacks poem level metrics (i.e. each verse is considered individually and not as a part of a whole). They base their evaluation on comparing generated poetry of different groups based on how the genetic algorithm was used. They use very broad questions such as \textit{which poem is more creative} or \textit{which poem has better imagery}. This is potentially problematic as broad questions open more room for subjective interpretation. 

Last year, a myriad of work on generation of Chinese poetry with machine learning methods was conducted. Research ranging from mutual reinforcement learning \cite{yi-etal-2018-automatic} and conditional variational autoencoders \cite{li-etal-2018-generating-classical} to sequence-to-sequence Bi-LSTMs \cite{yang-etal-2018-stylistic} was presented. However, none of these methods has been motivated from the point of view of computational creativity, but rather serve for a purely generative purpose.

The work conducted by \citet{colton2012full}, although not recent, deserves special attention, as they had used the same FACE model as a basis in their poem generation. They take a template based approach to generating poems from current news articles. Unfortunately they do not provide an evaluation of the generated poetry, which makes meaningful comparison difficult.

The work presented by us in this paper has to deal with the rich morphosyntax of Finnish, which is an NLG problem far from solved. \citet{poem_gen} presents a solution for this problem in their Finnish poem generator. However, their generator relies on predefined rule-based structures, whereas our aim is to have a system with more structural versatility, and yet the capability of coping with the morphosyntax.

\section{Creativity}

In order to separate our system from generative non-creative systems, we have to provide some justification as to why our system would exhibit creativity in the first place. For this reason, we follow the SPECS approach \cite{Jordanous2012} that has been designed to evaluate creativity in a reasoned fashion. The approach requires creativity to be defined first on an abstract level, and then, following the abstract definition, creativity should be defined in the context of the creative task that is to be solved. After establishing these definitions, creativity of the system should then be evaluated based on the definitions.

\subsection{Creativity in General}

For an abstract level definition of creativity, we use FACE \cite{colton2011computational}. The theory divides creative act into two categories, one is the ground level generative act of producing an artefact and the other is on the level of the process. Both of these categories are represented in the four aspects of creativity: framing, aesthetics, concept and expression.

Framing consists of outputting a framing for a creative artefact, and the process that generates this output. The framing should be an explanation in natural language, for instance, putting the created artefact into a historical and cultural context or describing the processes of creating the output artefact. In other words, framing can be used as an additional persuasive or explanatory message that is delivered to the human perceiving the artefact produced by a computational system.

Aesthetics consist of a function measuring the aesthetic quality of the output and/or the program producing it. On the process level, FACE takes into account how the aesthetic measures came to be in the system. The system should be able to assess its own work and rate its creations. This aesthetic measure can also be used to computationally assess artefacts produced by other systems or humans.

Concept is used to refer to the program that generates creative artefacts on the ground level. And on the process level it refers to how such a program was generated. Finally, the ground level expression is the creative output, or artefact, generated by the system, whereas the process level of expression describes the method for generating output for a given input.

\subsection{Creativity for Our Poem Generator}

As framing can exists in many different forms according to the original FACE model, we follow a more narrowed down notion of framing, which is the intention of the computer in creating artefacts \cite{charnley2012notion}. In other words, the computer should be able to output a justification explaining what certain aspects of the poem mean. The importance of framing has recently been highlighted in the literature \cite{framing2019}. 

Framing does not have to be a creative act on its own. In our case, the process of coming up with a framing is a template based approach that conveys the intent of the creative program in producing the output poem. This intent, on the other hand is captured by the aesthetic function of the creative system. Therefore, the framing produced should explain the poem in terms of the aesthetic measures.

Poetry as a genre showcases a wide diversity in terms of aesthetics; ranging from epic poetry following a strict meter to modern free form poetry. Even to a degree that the poetic genre has become fragmented ever since the 20th century \cite{juntunen_2012}. This diversity is not just limited to the level of structure, but is also reflected in meaning - some forms of poetry are meant to be read and interpreted literally, where as others rely on indirect communication such as symbolism and metaphors (see \citealp{runoousoppi}). In our work, we are not aiming to model the poetic genre as a whole, but rather define a set of aesthetic functions that capture different aspects in poetry ranging from the structural to the meaning.

In terms of structure, our system should be able to assess rhyming in its various forms (alliteration, assonance, consonance and full rhymes) and the meter of the poetry as defined by poetic foot and syllable count.

For meaning, our system should be able appreciate the presence of metaphors, the semantic coherence of the words forming the poem and in especial the presence of words forming different semantic fields and the semantic difference of these fields as an indicator of tension built by the choice of words in a poem (cf. \citealp{lotman}).

Certain poems paint a mental image in the mind of the reader; this qualia\footnote{For more on the problem of qualia, see \cite{chalmers1995absent}} provoking aspect of poetry is called \textit{imagery}. As it is extremely difficult for a computer to assess such rich mental sensory phenomena provoked by poetry in humans, we have to reduce the aesthetics related to imagery to a more computationally manageable level, namely that of sentiment. Sentiments expressed in a poem can be indicators of the potential mood evoked by the sensory imagery in the poem. Another indicator of imagery is the use of concrete expressions (see \citealp{concretepoems}).

Although the list of aesthetic measures is predefined, from the point of view of the process, our system should be able learn to adjust its aesthetic measures based on existing poetry. Furthermore, we aim towards a system that can learn aesthetics of its own on its own level of abstraction, hence the use of apprentice.

In our case, the system consists of two concepts. One of them is a genetic algorithm (master) that has been defined by us, the programmers. The role of the master is to produce expressions through a search informed by the aesthetic functions. These expression are used to train the second concept, which is a sequence-to-sequence BRNN model (apprentice). This way, the overall system is given the capability of producing new concepts of its own.

The expressions output by the system are computationally created Finnish poems. Ultimately, we evaluate the expressions produced by the apprentice with real humans and by the master's aesthetic measures.

\subsection{Data}

We use the 6,189 Finnish poems that are available on Wikisources\footnote{https://fi.wikisource.org} as our poem corpus. We use the Finnish dependency parser \cite{Haverinen2014} to parse the poems for morphological features, syntactic relations, part of speech and lemma for each word. The parsing is done on a verse-level. We split each poem into stanzas as divided in Wikisources. From now on we refer to a stanza of an existing poem simply as a poem. The reason for this is to have shorter poems to deal with in the generation step. This is especially important for the human evaluation as shorter poems can be evaluated more accurately, as longer poems have more room for unintentional characteristics that can be interpreted too positively by human judges, such as a perceivably deeper meaning that is due to the mere fact of having more context to read more into. After splitting the poems into stanzas, we have a total of 34,988 poems.

We use the word embeddings\footnote{http://bionlp-www.utu.fi/fin-vector-space-models/fin-word2vec-lemma.bin} that have been trained on the Finnish Internet Parsebank \cite{laippala2014syntactic}. We prefer this model for two reasons: first it has been trained on a 1.5 billion token corpus that is big on the Finnish scale and second it has been trained on lemmas, which is an important factor for a highly agglutinating language such as Finnish. In order to generate grammatical Finnish, the words need to be inflected. This step is easier if the replacement words are already in a lemmatized from.

\section{Generating Poetry}

The master-apprentice approach outlined in \citet{inlg} consists of a creative master, which is a genetic algorithm, and an apprentice, which is a sequence-to-sequence model. In this part of the paper, we describe how the aesthetics are implemented in the master and how it is used to generate poems for the apprentice to learn from.

In this paper, we experiment with two different masters, which will learn the weights for their aesthetic functions form poems of different eras. We use these masters to train one apprentice for each of them. In addition, we train one apprentice, which will learn from both of the masters.

\subsection{Master}

The master is a genetic algorithm following the implementation presented in \citet{alnajjar2018slogans}. In practice, the algorithm takes in a random poem from the poem corpus and uses it to produce an initial population of 100 individuals. These individuals produce an offspring of another 100 individuals that go through mutation and crossover, and at the end of each generation the individuals are scored according to the aesthetic functions defined later in this section. The 100 fittest individuals are selected with NSGA-II algorithm \cite{Deb:2002:FEM:2221359.2221582} to survive to the next generation. This process is done for 50 generations.

All individuals in the initial population are based on a randomly selected poem and a randomly picked theme word. The theme is expanded into the 30 most semantically similar words to the theme word using word2vec \cite{mikolov2013distributed}. Each poem in the initial population is assigned a random theme out of the 30 semantically similar words to the theme. Additionally, we modify each poem in the initial population once by using the mutation function. This is applied to have more variety of poems in the initial population given that all of them are based on the same original poem from the corpus.

In mutation, a random content word is picked in the poem and it is replaced by a word related to the input theme (assigned to the poem) or by a word that is similar to the original one, while ensuring that the new replacement matches the original in terms of its part-of-speech. To obtain words that are related to the input theme, we build a semantic relatedness model following \citet{meta4meaning981} using the flat 5-gram data provided by \citet{laippala2014syntactic} as the corpus.
Regarding the semantic similarity to the original word, we utilize the word2vec word embeddings model. The space of candidate replacements consists of the top 1,000 and 300 (empirically chosen) semantically related and similar words, respectively. Out of these candidates, only words that match the part-of-speech of the original word, based on UralicNLP \cite{uralicnlp_2019}, are considered in the random selection.

In terms of the crossover, we employ a single-point crossover on a verse-level where one point in both individuals is selected at random and verses to the right of that point are swapped.

As mutations and crossovers are bound to break the morphosyntax of Finnish, the new words are always inflected to match the original morphology with UralicNLP and Omorfi \cite{omorfi}. This will account for morphological agreement, but not for case government. In case government, the case of the complements of the verb depends on the verb itself. For this reason, we inflect words with an object relation with Syntax Maker \cite{hamalainen2018development} to produce a grammatical surface form even if the predicate verb is changed.

\subsubsection{Aesthetics}

To assess the sonic structure of poetry the following rule-based aesthetic functions are defined on an inter-verse level: full rhyme, assonance and consonance. These count the number of rhyming words between verses of the poem. Alliteration is a metric calculated within a verse, as this type of rhyming occurs typically inside of a verse in Finnish poetry. As Finnish spelling is almost one to one mapping with phonology, we can do this on a character level without the need to approximate the pronunciation.

Meter is captured by two aesthetic functions: the number of syllables and the distribution of long and short syllables within a verse. These two functions are again solved by simple rules. The master rates higher the meter it has learned from its training corpus.

A previous attempt to capture imagery in the literature is by comparing the number of abstract and non-abstract words with the hypothesis that non-abstract words provoke more mental imagery \cite{kao2012computational}. However, this notion can be used only as a proxy to the quantity of imagery in poetry, but it tells nothing about the nature of the provoked imagery. For this reason, we have also decided to use sentiment as an indicator of the mood of the mental image painted by the poem.

For abstractness of words we use an existing dataset for English that maps 40,000 common English words to an average concreteness score as annotated by humans on a 5-point Likert scale \cite{brysbaert2014concreteness}. We translate this data in Finnish with a Wiktionary based online dictionary\footnote{http://www.sanakirja.org/} in such a way that we consider the three top-most translations that are verbs, nouns or adjectives for each English word. To deal with polysemy, if multiple English words translate into one Finnish word, we take the average of the concreteness values of the English words for the Finnish word. If the concreteness value is greater or equal to 3, the word is considered concrete. The aesthetic function gives a ratio of concrete words over concrete and abstract words in the poem.

For sentiment, due to the lack of resources for Finnish, we use a recent state of the art method \cite{feng-wan-2019-learning} that can learn sentiment prediction for English with annotated data and use the model for other languages by bilingual word embeddings. We train the model with sentiment annotated data for English from the OpeNER project \cite{agerri2013opener}. We use their method to map the pretrained Finnish and English fasttext models from \citet{grave2018learning} into a common space. This aesthetic measure will score sentiments on verse level and output their variance on the poem level.

Dividing words into semantic fields can be used as an auxiliary tool in poem analysis in literature studies as it can reveal tensions inside of a poem (c.f \citealp{lotman}). By following this notion, we cluster the open class part of speech words based on their cosine similarity within a poem. For this clustering, we use affinity propagation \cite{frey2007clustering}, which takes a similarity matrix as input and clusters the words based on the matrix. The number of clusters is not fixed and affinity propagation is free to divide the words in as many clusters as necessary.

The clustering aesthetic function looks at the number of clusters in a poem and the average semantic distances of the clusters. The distance between two clusters is calculated by counting a centroid for each cluster based on the word vectors of a cluster and then calculating the cosine distance of the centroids of the clusters. The values output by the aesthetic function will set standards to how semantically cohesive the words have to be with each other, and how distant can their meanings be.

Although words in different clusters might be distant \textit{semantically}, they can be related \textit{pragmatically}. Therefore, we want to reveal possible metaphorical interpretations of a given word in the poem. We represent each semantic cluster found in a poem by a single word. In doing so, we compute the centroid vector of words in each cluster and use the nearest word in the model's vocabulary to the centroid as the topic of the cluster. Thereafter, we iterate over all the possible combinations of having a certain topic as a tenor and another as a vehicle and measure the metaphoricity of the poem with respect to them. We measure that using the two metaphoriticy measurements defined by \citet{alnajjar2018slogans}, one for measuring how a word in the poem relates to both concepts and the other for measuring how related a word is to the vehicle but not to the tenor\footnote{See \cite{Richards36} for more on tenor and vehicle}. The metaphoricity value is then represented by the mean of the two measurements in case both had a positive value, otherwise zero is returned. Using the metaphoricity value assigned to each tenor-vehicle combination, we define two metaphoricity aesthetics 1) the maximum metaphoricity value and 2) the number of metaphorical clusters (i.e. combinations where the metaphoricity value is above zero).

As having many objectives is difficult in practice to handle for the NSGA-II algorithm (see \citealp{tanigaki2014preference}), we group the aesthetic functions into four fitness functions. Sonic (rhyme, alliteration, consonance, assonance, foot and syllable count), semantic (number of clusters and average and maximum distance between the clusters), imagerial (concrete word ratio and variance of sentiment) and metaphorical (the maximum score for metaphoricity and the number of metaphorical words) functions represent the four fitness functions used by the genetic algorithm. These fitness functions sum up the individual aesthetic functions when they are used to score a poem.

\subsection{Learning the Aesthetics}

We divide our corpus into centuries: the 19th and 20th century. We have two masters learn their aesthetics from either century making them specialized in that century in particular. We first learn weights for the individual aesthetic functions within the higher-level fitness function they belong to. We do this by training four random forest classifiers \cite{breiman2001random}, one for each of the four higher level fitness functions. The classifiers get the features produced by the aesthetic functions belonging to the fitness function in question. The classifiers are trained with the entire corpus to predict true for the desired century and false for other centuries. 

The trained classifiers are only used for their weights for each individual feature. These weights are used in the genetic algorithm to multiply the output of each aesthetic function adjusting their importance for the century.

As the weights tell only little about the possible values the aesthetic functions can or should have within one century, we calculate a range of accepted values for each aesthetic function within a century. The 25th percentile of the values is set as the minimum boundary of an accepted value and the 75th percentile as the maximum boundary. If the value output by the aesthetic function is outside of this range, the output value is set to 0.

\subsubsection{Master's liking}

For the evaluation purposes of the apprentices, it is important to set standards to what is good poetry according to the master. The master likes a poem generated by the apprentice if the poem gets a positive value in each one of the four fitness functions. If any of the values is 0, the master is considered not liking the poems.

\subsection{Apprentice}

Apprentice is a sequence-to-sequence model that learns to produce creatively altered verses out of verses in existing poetry. To achieve this, we use a BRNN model with a copy attention mechanism by using OpenNMT \cite{klein-etal-2018-opennmt}. We use the default settings which are two layers for encoding and decoding and general global attention \cite{luong2015effective}.

One apprentice is trained from the output of each master, and an additional one from the output of both of the masters. We train the apprentices for 90000 steps to produce poems one verse at a time, from the original poem to the master generated ones. The master for the 19th century produced 11903 poems and the 20th century one 11900 poems out of randomly picked initial poems from the entire corpus. These constitute the training data for the apprentices. The random seed used in training is the same for all apprentices to make intercomparison possible.

\section{Results and Evaluation}

Evaluation is one of the most important and difficult parts of computational creativity, however it is oftentimes overlooked and conducted in an ad-hoc manner with little to do with the actual problem being modelled \cite{lamb2018evaluating}. In practice this means that a great deal of work is evaluated based on questions and metrics that have not been justified. This practice together with the issue expressed by \citet{veale2016shape} that people are ready to read more into the output if it has a suitable linguistic structure regardless of the actual underlying creative intent of the system, are things that should not go unnoticed when evaluating a computationally creative system.

\begin{quote}
\small
    \begin{tabular}{ll}
        \textit{Mutta hyökkäykset, jotka kestää sain,}  \\
        \textit{muistot, jotka rakkauden estää,}  \\
        \textit{esiin ilmentyy vihaa kasvattain.}  \\ 
        \\
        But the attacks I was to endure, \\
        the memories that prevent love, \\
        emerge amplifying the ire \\
    \end{tabular}
\end{quote}

Above is an example poem output by the master in Finnish followed by its translation in English. The example is of a typical length of a poem produced by the system as the human authored poems were split into stanzas.

\subsection{Concepts}

The master as a concept is fixed and can only adjust its appreciation, but the apprentice is an entirely new concept that is created from the output of the master. In this section we evaluate the apprentices by evaluating their output by masters' liking. This is done only in an automatic fashion by having all 3 of the apprentices create output for 100 randomly picked poems from the poem corpus. 

\begin{table}[ht!]
\small
\centering
\begin{tabular}{|l|c|c|}
\hline
 & master 1800 & master 1900 \\ \hline
apprentice 1800 & 28\% & 33\% \\ \hline
apprentice 1900 & 36\% & 39\% \\ \hline
apprentice both & 47\% & 51\% \\ \hline
\end{tabular}
\caption{The percentage of the output of the two masters liked}
\label{tab:concepts}
\end{table}

Table \ref{tab:concepts} shows how many of the poems produced by the different apprentices the masters liked. It is clear from the results that the apprentices did not do too well in terms of learning the century specific aesthetics. Nevertheless, having both of the centuries in the training boosted the results in terms of the two masters liking the poems. This is probably due to the fact of having more training data available.

\subsection{Framing and Aesthetics}

In order to make it less likely that people read more into the poems than what is there, we evaluate the poems with people based on the framing produced by the system. The main purpose of this evaluation is not to evaluate how \textit{good} the output poems are, but how often the aesthetic functions agree with human judgment. The framing consists of templates that the system fills based on its aesthetic functions. People are asked whether they agree or disagree with the statements expressed in the framing. In addition, people have the possibility of stating that they don't know whether to agree or disagree.

For the evaluation, we sampled 30 poems at random from the poetry generated by the two masters. We printed each poem 5 times, and we divided each set of 30 unique poems into 3 piles of 10 poems with their framing. Each pile was shuffled so that no pile contained exactly the same poems and no pile had the same order for the poems. The shuffling was done to decrease any potential bias introduced by the order of presentation of the poems.

Initially, we recruited 15 people, each one to go through one pile of 10 poems. However, 5 people found the task too time consuming and stopped after evaluating a few poems. The unevaluated poems from these piles were assigned to completely new reviewers. In the end, each unique poem was evaluated 5 times by different people and no individual evaluator evaluated more than 10 poems.

A framing was generated for each poem. The framing followed always the same structure. The first statements relating to rhyming were presented as questions whereas the rest of them were statements. The statements were formed in the following way (translated from Finnish):

\begin{enumerate}
  \itemsep0em
  \item Do the words written in italics have rhymes (e.g. heikko peikko)?
  \item Do the words written in italics have assonance (e.g. t\textbf{a}l\textbf{o} s\textbf{a}n\textbf{o})?
  \item Do the words written in italics have consonance (e.g. \textbf{s}a\textbf{kk}o \textbf{s}o\textbf{kk}a)?
  \item Does the poem have alliteration within a verse (e.g. \textbf{v}anha \textbf{v}esi)?
  \item Verse number X and Y have the same meter
  \item The poem has X semantic fields: [semantic cluster 1]... and [semantic cluster N]
  \item The semantic fields [semantic cluster X] and [semantic cluster Y] are the closest to each other
  \item The semantic fields [semantic cluster A] and [semantic cluster B] are the furthest away from each other
  \item The following words in the poem [concrete words] are concrete concepts
  \item The verse number X is positive
  \item The verse number Y is negative
  \item The following words in the poem [metaphorical words] can be understood metaphorically
  \item The word X has a metaphorical connection to word Y
\end{enumerate}

For the questions on rhyming, the system highlights in italics all the words that have one of the rhyming types. For the meter statement and negative and positive verse statements, random numbers are picked within the range of the length of the poem. For these questions, people agreeing does not produce the highest score, but rather if people's prediction is in line with the prediction of the aesthetic function. Also, if the poem didn't have any metaphorical words, random words were picked for the last two questions. Again, if people disagreed when random words were presented and agreed when actual metaphorical words were presented, the accuracy of the system based on the evaluation would go higher.

\begin{figure}[!htb]
\center{\includegraphics[width=8cm]
{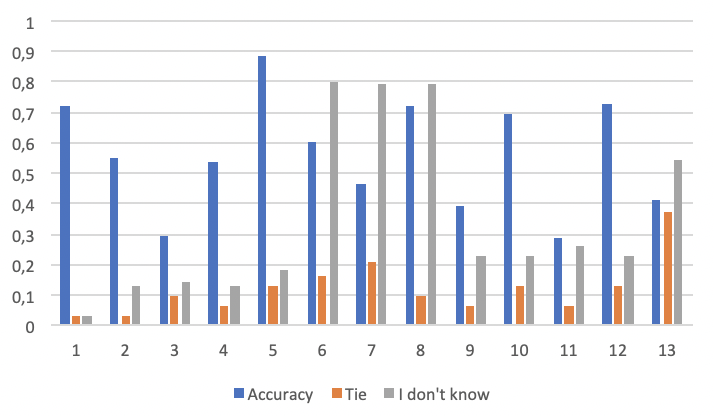}}
\caption{\label{fig:results} Evaluation results for aesthetics and framing}
\end{figure}

The accuracy reported in Figure \ref{fig:results} shows how often the prediction (agree/disagree) of the aesthetic functions matches that of the majority of the people out of all the times a majority decision could be reached per poem. The tie shown in the figure shows the percentage of time the statement received an equal number of agreeing and disagreeing opinions from people per poem. The data show by \textit{I don't know} represents the number of times people stated they did not know over all the answers for the statement. Note that this is not calculated per poem but per statement.

The statements related to semantics were the most difficult ones for people to evaluate with around 80\% of the time people saying they did not know whether to agree or not. Another difficult statement to judge was the last metaphorical statement including an interpretation for two words being metaphorically connected. This question also included the highest number of ties in people’s judgments.

Interestingly, the accuracy was high only for the traditional rhyme types, but lower on the assonance and consonance. Even though our rules can easily and objectively measure the existence of these rhyming types, it is interesting to see that people's judgment deviates from the values output by the aesthetic functions. Especially revealing is the low accuracy on consonance. Our system sees consonance whenever two words have the same consonants in the same positions such as in \textit{\underline{j}o} (already) and \textit{\underline{j}a} (and) or \textit{e\underline{n}} (I don't) and \textit{o\underline{n}} (is). Even though these words do exhibit consonance, it seems that people do not find such consonance \textit{perceivable}. This being said, the mere existence of rhyming is not enough, but it should also be perceivable. Just what this perceivability entails is an interesting question left for future research.

For semantics it is difficult to draw any meaningful conclusions as more often than not, people simply did not know whether to agree or not. However, the results do seem promising for the correctness semantic clusters (60 \% of the time) and the furthest clusters (72\% of the time). At any rate, semantics calls for further qualitative analysis in the future as it seems to be a difficult thing to assess for people.

In the case of imagery, it seems that people agreed on the concreteness 39\% of the time, although the score might seem low, it is to remember that all the concrete words were presented as a list in the framing. If even one of the words was not perceived as concrete, people were likely to disagree. Sentiment, on the other hand, resulted in mixed accuracies; the accuracy for positive sentiment was 69\% whereas for negative sentiment the accuracy was 28\%. As the sentiment analysis was based on an existing state-of-the-art method, this result is surprising. However, it is very likely the case that negativity in poetry is expressed in a very different way than in other text types. In other words, there is a need for a sentiment annotated corpus consisting of poetry and other literary texts for better predicting the sentiment in poems. All in all, the prediction for concrete words could also benefit from a dataset authored specifically for Finnish.

The accuracy for the metaphorical words was high, 73\%. However, the interpretation provided for one of the metaphorical words gave inconclusive results, as people either did not know or had very mixed judgments. This part as well calls for qualitative analysis in the future.

\subsection{Expressions}

Finally we evaluate the expressions of the master and the apprentice in relation to each other. For this evaluation we treat both of the masters as one, and we evaluate the best apprentice according to the masters' liking. We sample randomly 10 poems from the corpus for which both the master and the apprentice had produced altered poems. We evaluate these poems by asking people which one of the generated poems from the same original one they prefer, that of the master or that of the apprentice. We present the two poems on the same page, shuffling their order for each printout. We also shuffle the order of the poems. We ask 10 people to rate the 20 poems, 10 master generated and 10 apprentice generated ones.

\begin{figure}[!htb]
\center{\includegraphics[width=7cm]
{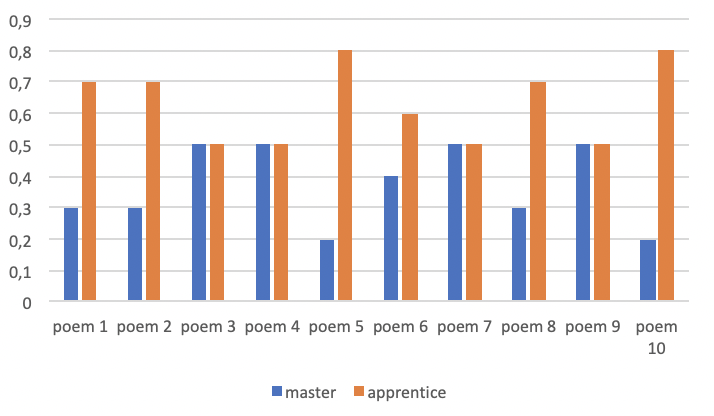}}
\caption{\label{fig:results2} People's preference for each poem}
\end{figure}

Figure \ref{fig:results2} shows the preference of the people per poem. The poetry generated by the apprentice was most often preferred by the judges. The master generated poetry did not reach to a majority in preference for any poem. The interesting question of what happens in the poems that result in a tie in people's preference calls for a future qualitative study to understand better the phenomenon of the evaluation.

\section{Conclusions}

We have shown our novel method for generating poetry in Finnish. With the help of the FACE model, we were able to conduct evaluation on the aesthetics and framing that was revealing of the shortcomings of our system. Framing made it possible to assess the core functionality better by minimizing the room for people reading more into the poem than what was there. Having the option for people to say that they do not know rather than forcing them to either agree or disagree revealed the difficulty of assessing semantics and metaphors even for people. We propose for the future to conduct evaluation on such high level features of language on a qualitative fashion to better understand how people perceive these in generated poetry.

As a vast majority of the NLP research focuses on English, we had to deal with the practical issue of the scarce annotated resources for Finnish to capture the high level features such as concreteness, sentiment and metaphor. As a result we ended up developing useful resources for the aesthetic functions which we have made publicly available on Github\footnote{https://github.com/mikahama/finmeter}.

\bibliography{acl2019}

\begin{thebibliography}{44}
\expandafter\ifx\csname natexlab\endcsname\relax\def\natexlab#1{#1}\fi

\bibitem[{Agerri et~al.(2013)Agerri, Cuadros, Gaines, and
  Rigau}]{agerri2013opener}
Rodrigo Agerri, Montse Cuadros, Sean Gaines, and German Rigau. 2013.
\newblock Opener: Open polarity enhanced named entity recognition.
\newblock \emph{Procesamiento del Lenguaje Natural}, (51).

\bibitem[{Alnajjar et~al.(2018)Alnajjar, Hadaytullah, and
  Toivonen}]{alnajjar2018slogans}
Khalid Alnajjar, Hadaytullah Hadaytullah, and Hannu Toivonen. 2018.
\newblock {``Talent, Skill and Support.'' A} method for automatic creation of
  slogans.
\newblock In \emph{Proceedings of the 9th International Conference on
  Computational Creativity (ICCC 2018)}, pages 88--95, Salamanca, Spain.
  Association for Computational Creativity.

\bibitem[{Alnajjar and H{\"a}m{\"a}l{\"a}inen(2018)}]{inlg}
Khalid Alnajjar and Mika H{\"a}m{\"a}l{\"a}inen. 2018.
\newblock A master-apprentice approach to automatic creation of culturally
  satirical movie titles.
\newblock In \emph{Proceedings of the 11th International Conference on Natural
  Language Generation}, pages 274--283.

\bibitem[{Breiman(2001)}]{breiman2001random}
Leo Breiman. 2001.
\newblock Random forests.
\newblock \emph{Machine learning}, 45(1):5--32.

\bibitem[{Brysbaert et~al.(2014)Brysbaert, Warriner, and
  Kuperman}]{brysbaert2014concreteness}
Marc Brysbaert, Amy~Beth Warriner, and Victor Kuperman. 2014.
\newblock Concreteness ratings for 40 thousand generally known english word
  lemmas.
\newblock \emph{Behavior research methods}, 46(3):904--911.

\bibitem[{Burroway(2007)}]{concretepoems}
Janet Burroway. 2007.
\newblock \emph{Imaginative Writing: The Elements of Craft}.
\newblock Pearson.

\bibitem[{Chalmers(1995)}]{chalmers1995absent}
David~J Chalmers. 1995.
\newblock Absent qualia, fading qualia, dancing qualia.
\newblock \emph{Conscious experience}, pages 309--328.

\bibitem[{Charnley et~al.(2012)Charnley, Pease, and
  Colton}]{charnley2012notion}
John~William Charnley, Alison Pease, and Simon Colton. 2012.
\newblock On the notion of framing in computational creativity.
\newblock In \emph{ICCC}, pages 77--81.

\bibitem[{Colton et~al.(2011)Colton, Charnley, and
  Pease}]{colton2011computational}
Simon Colton, John~William Charnley, and Alison Pease. 2011.
\newblock Computational creativity theory: The {FACE} and {IDEA} descriptive
  models.
\newblock In \emph{ICCC}, pages 90--95.

\bibitem[{Colton et~al.(2012)Colton, Goodwin, and Veale}]{colton2012full}
Simon Colton, Jacob Goodwin, and Tony Veale. 2012.
\newblock Full-{FACE} poetry generation.
\newblock In \emph{ICCC}, pages 95--102.

\bibitem[{Cook et~al.(2019)Cook, Colton, Pease, and Llano}]{framing2019}
Michael Cook, Simon Colton, Alison Pease, and Maria~Theresa Llano. 2019.
\newblock Framing in computational creativity – a survey and taxonomy.
\newblock In \emph{The proceedings of the tenth international conference on
  computational creativity}, pages 156--163.

\bibitem[{Deb et~al.(2002)Deb, Pratap, Agarwal, and
  Meyarivan}]{Deb:2002:FEM:2221359.2221582}
K.~Deb, A.~Pratap, S.~Agarwal, and T.~Meyarivan. 2002.
\newblock \href {https://doi.org/10.1109/4235.996017} {A fast and elitist
  multiobjective genetic algorithm: Nsga-ii}.
\newblock \emph{Trans. Evol. Comp}, 6(2):182--197.

\bibitem[{Feng and Wan(2019)}]{feng-wan-2019-learning}
Yanlin Feng and Xiaojun Wan. 2019.
\newblock \href {https://www.aclweb.org/anthology/N19-1040} {Learning bilingual
  sentiment-specific word embeddings without cross-lingual supervision}.
\newblock In \emph{Proceedings of the 2019 Conference of the North {A}merican
  Chapter of the Association for Computational Linguistics: Human Language
  Technologies, Volume 1 (Long and Short Papers)}, pages 420--429, Minneapolis,
  Minnesota. Association for Computational Linguistics.

\bibitem[{Frey and Dueck(2007)}]{frey2007clustering}
Brendan~J Frey and Delbert Dueck. 2007.
\newblock Clustering by passing messages between data points.
\newblock \emph{science}, 315(5814):972--976.

\bibitem[{Gerv{\'a}s(2001)}]{gervas2001expert}
Pablo Gerv{\'a}s. 2001.
\newblock An expert system for the composition of formal {S}panish poetry.
\newblock In \emph{Applications and Innovations in Intelligent Systems VIII},
  pages 19--32. Springer.

\bibitem[{Grave et~al.(2018)Grave, Bojanowski, Gupta, Joulin, and
  Mikolov}]{grave2018learning}
Edouard Grave, Piotr Bojanowski, Prakhar Gupta, Armand Joulin, and Tomas
  Mikolov. 2018.
\newblock Learning word vectors for 157 languages.
\newblock In \emph{Proceedings of the International Conference on Language
  Resources and Evaluation (LREC 2018)}.

\bibitem[{H{\"a}m{\"a}l{\"a}inen(2018)}]{poem_gen}
Mika H{\"a}m{\"a}l{\"a}inen. 2018.
\newblock Harnessing {NLG} to create {F}innish poetry automatically.
\newblock In \emph{Proceedings of the Ninth International Conference on
  Computational Creativity}, pages {9--15}.

\bibitem[{H{\"a}m{\"a}l{\"a}inen(2019)}]{uralicnlp_2019}
Mika H{\"a}m{\"a}l{\"a}inen. 2019.
\newblock \href {https://doi.org/10.21105/joss.01345} {{UralicNLP}: An {NLP}
  library for {U}ralic languages}.
\newblock \emph{Journal of Open Source Software}, 4(37):1345.

\bibitem[{H{\"a}m{\"a}l{\"a}inen and Rueter(2018)}]{hamalainen2018development}
Mika H{\"a}m{\"a}l{\"a}inen and Jack Rueter. 2018.
\newblock {Development of an Open Source Natural Language Generation Tool for
  Finnish}.
\newblock In \emph{Proceedings of the Fourth International Workshop on
  Computational Linguistics for {U}ralic Languages}, pages 51--58.

\bibitem[{Haverinen et~al.(2014)Haverinen, Nyblom, Viljanen, Laippala, Kohonen,
  Missil{\"a}, Ojala, Salakoski, and Ginter}]{Haverinen2014}
Katri Haverinen, Jenna Nyblom, Timo Viljanen, Veronika Laippala, Samuel
  Kohonen, Anna Missil{\"a}, Stina Ojala, Tapio Salakoski, and Filip Ginter.
  2014.
\newblock \href {https://doi.org/10.1007/s10579-013-9244-1} {Building the
  essential resources for {F}innish: the {T}urku dependency treebank}.
\newblock \emph{Language Resources and Evaluation}, 48(3):493--531.

\bibitem[{Jennings(2010)}]{Jennings2010}
Kyle~E. Jennings. 2010.
\newblock \href {https://doi.org/10.1007/s11023-010-9206-y} {{Developing
  Creativity: Artificial Barriers in Artificial Intelligence}}.
\newblock \emph{Minds and Machines}, 20(4):489--501.

\bibitem[{Jordanous(2012)}]{Jordanous2012}
Anna Jordanous. 2012.
\newblock \href {https://doi.org/10.1007/s12559-012-9156-1} {A standardised
  procedure for evaluating creative systems: Computational creativity
  evaluation based on what it is to be creative}.
\newblock \emph{Cognitive Computation}, 4(3):246--279.

\bibitem[{Juntunen(2012)}]{juntunen_2012}
Tuomas Juntunen. 2012.
\newblock {Kirjallisuudentutkimus}.
\newblock In \emph{Genreanalyysi: tekstilajitutkimuksen k{\"a}sikirja}, pages
  528–--536.

\bibitem[{Kanerva et~al.(2014)Kanerva, Luotolahti, Laippala, and
  Ginter}]{laippala2014syntactic}
Jenna Kanerva, Juhani Luotolahti, Veronika Laippala, and Filip Ginter. 2014.
\newblock Syntactic n-gram collection from a large-scale corpus of internet
  {F}innish.
\newblock In \emph{Human Language Technologies-The Baltic Perspective:
  Proceedings of the Sixth International Conference Baltic HLT}, volume 268,
  pages 184--191.

\bibitem[{Kantokorpi et~al.(1990)Kantokorpi, Pirjo, and Auli}]{runoousoppi}
Mervi Kantokorpi, Lyytikäinen Pirjo, and Viikari Auli. 1990.
\newblock \emph{Runousopin perusteet}.
\newblock Gaudeamus.

\bibitem[{Kao and Jurafsky(2012)}]{kao2012computational}
Justine Kao and Dan Jurafsky. 2012.
\newblock A computational analysis of style, affect, and imagery in
  contemporary poetry.
\newblock In \emph{Proceedings of the NAACL-HLT 2012 Workshop on Computational
  Linguistics for Literature}, pages 8--17.

\bibitem[{Klein et~al.(2018)Klein, Kim, Deng, Nguyen, Senellart, and
  Rush}]{klein-etal-2018-opennmt}
Guillaume Klein, Yoon Kim, Yuntian Deng, Vincent Nguyen, Jean Senellart, and
  Alexander Rush. 2018.
\newblock \href {https://www.aclweb.org/anthology/W18-1817} {{O}pen{NMT}:
  Neural machine translation toolkit}.
\newblock In \emph{Proceedings of the 13th Conference of the Association for
  Machine Translation in the {A}mericas (Volume 1: Research Papers)}, pages
  177--184, Boston, MA. Association for Machine Translation in the Americas.

\bibitem[{Lamb and Brown(2019)}]{twitsong3}
Carolyn Lamb and Daniel~G. Brown. 2019.
\newblock {TwitSong} 3.0: towards semantic revisions in computational poetry.
\newblock In \emph{Proceedings of the Tenth International Conference on
  Computational Creativity}, pages 212--219.

\bibitem[{Lamb et~al.(2018)Lamb, Brown, and Clarke}]{lamb2018evaluating}
Carolyn Lamb, Daniel~G Brown, and Charles~LA Clarke. 2018.
\newblock Evaluating computational creativity: An interdisciplinary tutorial.
\newblock \emph{ACM Computing Surveys (CSUR)}, 51(2):28.

\bibitem[{Li et~al.(2018)Li, Song, Zhang, Chen, Shi, Zhao, and
  Yan}]{li-etal-2018-generating-classical}
Juntao Li, Yan Song, Haisong Zhang, Dongmin Chen, Shuming Shi, Dongyan Zhao,
  and Rui Yan. 2018.
\newblock \href {https://www.aclweb.org/anthology/D18-1423} {Generating
  classical {C}hinese poems via conditional variational autoencoder and
  adversarial training}.
\newblock In \emph{Proceedings of the 2018 Conference on Empirical Methods in
  Natural Language Processing}, pages 3890--3900, Brussels, Belgium.
  Association for Computational Linguistics.

\bibitem[{Lotman(1974)}]{lotman}
Juri Lotman. 1974.
\newblock \emph{Den poetiska texten}.
\newblock Stockholm.

\bibitem[{Luong et~al.(2015)Luong, Pham, and Manning}]{luong2015effective}
Minh-Thang Luong, Hieu Pham, and Christopher~D Manning. 2015.
\newblock Effective approaches to attention-based neural machine translation.
\newblock \emph{arXiv preprint arXiv:1508.04025}.

\bibitem[{Mikolov et~al.(2013)Mikolov, Sutskever, Chen, Corrado, and
  Dean}]{mikolov2013distributed}
Tomas Mikolov, Ilya Sutskever, Kai Chen, Greg~S Corrado, and Jeff Dean. 2013.
\newblock Distributed representations of words and phrases and their
  compositionality.
\newblock In \emph{Advances in neural information processing systems}, pages
  3111--3119.

\bibitem[{Misztal and Indurkhya(2014)}]{misztal2014poetry}
Joanna Misztal and Bipin Indurkhya. 2014.
\newblock Poetry generation system with an emotional personality.
\newblock In \emph{ICCC}, pages 72--81.

\bibitem[{Oliveira(2017)}]{goncalo-oliveira-2017-survey}
Hugo~Gon{\c{c}}alo Oliveira. 2017.
\newblock A survey on intelligent poetry generation: Languages, features,
  techniques, reutilisation and evaluation.
\newblock In \emph{Proceedings of the 10th International Conference on Natural
  Language Generation}, pages 11--20, Santiago de Compostela, Spain.
  Association for Computational Linguistics.

\bibitem[{Oliveira et~al.(2017)Oliveira, Herv{\'a}s, D{\'\i}az, and
  Gerv{\'a}s}]{oliveira2017multilingual}
Hugo~Gon{\c{c}}alo Oliveira, Raquel Herv{\'a}s, Alberto D{\'\i}az, and Pablo
  Gerv{\'a}s. 2017.
\newblock Multilingual extension and evaluation of a poetry generator.
\newblock \emph{Natural Language Engineering}, 23(6):929--967.

\bibitem[{Pirinen et~al.(2017)Pirinen, Listenmaa, Johnson, Tyers, and
  Kuokkala}]{omorfi}
Tommi~A Pirinen, Inari Listenmaa, Ryan Johnson, Francis~M. Tyers, and Juha
  Kuokkala. 2017.
\newblock \href {http://hdl.handle.net/11372/LRT-1992} {Open morphology of
  finnish}.
\newblock {LINDAT}/{CLARIN} digital library at the Institute of Formal and
  Applied Linguistics, Charles University.

\bibitem[{Richards(1936)}]{Richards36}
Ivor~Armstrong Richards. 1936.
\newblock \emph{The Philosophy of Rhetoric}.
\newblock Oxford University Press, London, United Kingdom.

\bibitem[{Tanigaki et~al.(2014)Tanigaki, Narukawa, Nojima, and
  Ishibuch}]{tanigaki2014preference}
Yuki Tanigaki, Kaname Narukawa, Yusuke Nojima, and Hisao Ishibuch. 2014.
\newblock Preference-based nsga-ii for many-objective knapsack problems.
\newblock In \emph{2014 Joint 7th International Conference on Soft Computing
  and Intelligent Systems (SCIS) and 15th International Symposium on Advanced
  Intelligent Systems (ISIS)}, pages 637--642. IEEE.

\bibitem[{Toivanen et~al.(2012)Toivanen, Toivonen, Valitutti, and
  Gross}]{jukka}
Jukka Toivanen, Hannu Toivonen, Alessandro Valitutti, and Oskar Gross. 2012.
\newblock {Corpus-Based Generation of Content and Form in Poetry}.
\newblock In \emph{Proceedings of the Third International Conference on
  Computational Creativity}.

\bibitem[{Veale(2016)}]{veale2016shape}
Tony Veale. 2016.
\newblock The shape of tweets to come: automating language play in social
  networks.
\newblock \emph{Multiple Perspectives on Language Play. Mouton De Gruyter,
  Language Play and Creativity series}, pages 73--92.

\bibitem[{Xiao et~al.(2016)Xiao, Alnajjar, Granroth-Wilding, Agres, and
  Toivonen}]{meta4meaning981}
Ping Xiao, Khalid Alnajjar, Mark Granroth-Wilding, Kathleen Agres, and Hannu
  Toivonen. 2016.
\newblock {Meta4meaning}: Automatic metaphor interpretation using
  corpus-derived word associations.
\newblock In \emph{Proceedings of the 7th International Conference on
  Computational Creativity (ICCC 2016)}, Paris, France. Sony CSL, Sony CSL.

\bibitem[{Yang et~al.(2018)Yang, Sun, Yi, and Li}]{yang-etal-2018-stylistic}
Cheng Yang, Maosong Sun, Xiaoyuan Yi, and Wenhao Li. 2018.
\newblock \href {https://www.aclweb.org/anthology/D18-1430} {Stylistic
  {C}hinese poetry generation via unsupervised style disentanglement}.
\newblock In \emph{Proceedings of the 2018 Conference on Empirical Methods in
  Natural Language Processing}, pages 3960--3969, Brussels, Belgium.
  Association for Computational Linguistics.

\bibitem[{Yi et~al.(2018)Yi, Sun, Li, and Li}]{yi-etal-2018-automatic}
Xiaoyuan Yi, Maosong Sun, Ruoyu Li, and Wenhao Li. 2018.
\newblock \href {https://www.aclweb.org/anthology/D18-1353} {Automatic poetry
  generation with mutual reinforcement learning}.
\newblock In \emph{Proceedings of the 2018 Conference on Empirical Methods in
  Natural Language Processing}, pages 3143--3153, Brussels, Belgium.
  Association for Computational Linguistics.

\end{thebibliography}
\bibliographystyle{acl_natbib}

\end{document}